\documentclass[runningheads,orivec]{llncs}

\usepackage[T1]{fontenc}
\usepackage{graphicx}
\usepackage{booktabs}
\usepackage{multirow}
\usepackage{amsmath}
\usepackage{amssymb}
\usepackage{xspace}
\usepackage{microtype}
\usepackage[table,dvipsnames]{xcolor}
\usepackage{url}
\usepackage{tikz}
\usetikzlibrary{shapes.geometric, arrows.meta, positioning, fit, backgrounds, calc}
\usepackage[textsize=tiny]{todonotes}
\usepackage{hyperref}  
\usepackage{longtable}
\usepackage{graphicx}
\usepackage{subcaption}

\definecolor{lightgray}{gray}{0.92}
\definecolor{fmblue}{RGB}{220,235,250}
\definecolor{treegreen}{RGB}{220,245,225}

\newcommand{\highlightfirst}[1]{\textcolor{black}{#1}}
\newcommand{\highlightsecond}[1]{\textcolor{black}{#1}}

\begin{document}

\title{Tabular Foundation Models for Clinical Survival Analysis via Survival-Aware Adaptation \thanks{Accepted for publication at AIiH 2026.}
}

\titlerunning{Tabular Foundation Models For Clinical Survival Analysis}

\authorrunning{Minh-Khoi Pham et al.}

\author{Minh-Khoi Pham\inst{1,2}  \and
Luca Cotugno\inst{3} \and
Alina S\^irbu\inst{3} \and Tai Tan Mai\inst{1,2} \and Martin Crane\inst{1,2} \and Marija Bezbradica\inst{1,2}}
\institute{
ADAPT Centre, Dublin City University, Dublin, Ireland \and School of Computing, Dublin City University, Dublin, Ireland  \and
Department of Computer Science and Engineering, University of Bologna, Bologna, Italy\\
Corresponding author:
\email{minhkhoi.pham@adaptcentre.ie}
}

\maketitle

\begin{abstract}
Predicting time-to-event outcomes such as mortality is a fundamental task in clinical decision-making, commonly addressed through survival analysis. While classical statistical and deep learning approaches have been widely studied, they typically require task-specific training and sufficient labeled data. Recent advances in tabular foundation models offer a new paradigm by learning general-purpose representations for structured data. However, their applicability to censored time-to-event prediction in clinical settings remains underexplored, as typical applications are restricted to discrete classification rather than survival analysis tasks.  

In this work, we propose a lightweight adaptation approach for applying tabular foundation models to clinical survival analysis by directly training a survival-aware head on top of the pretrained representations. We study representative architectures, including TabPFN, TabDPT, and TabICL, and adapt them using a multi-task logistic regression (MTLR) head to model right-censored time-to-event outcomes. We evaluate this approach on a diverse set of public survival benchmarks and two large-scale ICU cohorts, MIMIC-IV and eICU.

Our results show that this transfer learning approach achieves competitive or superior performance compared to strong baselines. On MIMIC-IV, TabDPT-FT-MTLR reaches a C-index of $0.856$, corresponding to a relative improvement of +1.4\% over the best non-FM baseline (DeepSurv, $0.844$) and +6.7\% over the best zero-shot model ($0.802$). On eICU, TabICL-FT-MTLR achieves $0.797$, yielding gains of +1.7\% (DeepSurv, $0.784$) and +6.4\% ($0.749$), respectively.

These findings highlight the importance of combining pretrained tabular representations with survival-aware objectives and suggest that tabular foundation models provide a practical and effective alternative for clinical survival prediction. The code is publicly available at \url{https://github.com/kaylode/survival-fm}.
\end{abstract}


\section{Introduction}
\label{sec:introduction}

Predicting the time to clinically meaningful events, such as mortality, disease progression, or treatment failure, is a central problem in healthcare. These tasks fall under \textit{survival analysis}, which models time-to-event outcomes in the presence of censoring \cite{cox1972regression,kalbfleisch1980statistical}. In clinical practice, censoring is unavoidable: patients may be discharged, transferred, or remain event-free at the end of the observation period. Effective survival models must therefore learn from partially observed outcomes while providing reliable, individualized risk estimates over time.

Classical approaches, such as the Kaplan--Meier estimator and Cox proportional hazards (PH) model, remain widely used due to their interpretability and strong statistical foundations \cite{cox1972regression,kaplan1958nonparametric}. However, these methods rely on assumptions, including proportional hazards and linearity. These are frequently violated in modern clinical datasets, 
characterized by heterogeneous variables with complex interactions. Machine learning (ML) methods, including random survival forests and neural models such as DeepSurv and DeepHit, relax these assumptions and improve predictive flexibility \cite{katzman2018deepsurv,lee2018deephit}. Nevertheless, they typically require full-scale training, careful hyperparameter tuning, and substantial labeled data that are often difficult to satisfy in real-world clinical settings.

Recent advances in \textit{tabular foundation models} (FMs) offer a promising alternative for learning from structured clinical data. Models such as TabPFN \cite{hollmann2023} are pretrained on large collections of synthetic tabular tasks and can perform predictions via \textit{in-context learning}\footnote{In-context learning refers to making predictions by conditioning on a small set of labeled examples provided at inference time, without updating model parameters.}, eliminating the need for training on the target dataset. This approach enables strong generalizability in small-data regimes and simplifies deployment, making it very appealing for healthcare applications. 

Despite these advantages, tabular foundation models have been primarily developed for standard supervised tasks such as classification and regression, and their applicability to survival analysis remains largely unexplored. In particular, survival prediction introduces additional challenges, including censoring, time-dependent risk estimation, and evaluation under partial observability, which are not directly addressed by existing foundation model frameworks.

In this work, we investigate whether tabular foundation models can be effectively applied to clinical survival analysis. We study three representative architectures—TabPFN, TabDPT, and TabICL—and introduce parameter-efficient adaptation to enable time-to-event prediction under censoring. We evaluate these models on survival analysis tasks across multiple clinical datasets, comparing against established statistical and ML baselines.

Our results show that tabular foundation models, when combined with lightweight survival-aware adaptation, achieve competitive predictive performance while maintaining strong robustness across datasets and subpopulations. These findings suggest that tabular foundation models provide a practical and efficient alternative to traditional survival modeling approaches in clinical settings, without requiring extensive task-specific training.


\section{Related Work}

\paragraph{\textbf{Classical and machine learning survival models.}}
Survival analysis has been extensively studied in statistics and ML. Classical approaches such as the Kaplan--Meier estimator and Cox PH model remain widely used due to their interpretability and statistical grounding~\cite{cox1972regression,kaplan1958nonparametric}. Extensions including random survival forests and gradient boosting relax linearity assumptions and capture non-linear feature interactions, often achieving strong performance on structured clinical data~\cite{chen2013gradient,ishwaran2008}. However, these methods rely on hand-crafted features and may struggle to capture the complex dependencies in high-dimensional data.

\paragraph{\textbf{Deep learning for survival analysis.}}
Deep learning models extend classical approaches by learning representations directly from covariates. Cox-based neural models such as DeepSurv preserve the partial likelihood formulation, while methods such as DeepHit and MTLR enable more flexible modeling of non-proportional hazards and competing risks~\cite{katzman2018deepsurv,lee2018deephit,yu2011learning}. These approaches have shown improved performance in various clinical settings, including intensive care prediction tasks \cite{gomez2025benchmarking}. However, they typically require extensive end-to-end training, careful hyperparameter tuning, and sufficient labeled data, limiting their applicability in data-constrained environments.

\paragraph{\textbf{Tabular foundation models for clinical prediction.}}
Recent work on tabular foundation models introduces a new paradigm for learning from structured data ~\cite{hollmann2023,ma2024tabdpt,ye2024tabicl}. They are pretrained on large collections of synthetic tabular tasks and perform inference via in-context learning, enabling strong performance without gradient-based adaptation. These models are particularly effective in low-data regimes and offer simplified deployment compared to traditional ML pipelines. Recent work starts pivoting them into clinical domain \cite{pham2026retrieval}. However, they are primarily designed for classification and regression tasks and do not directly support censored time-to-event prediction.

\paragraph{\textbf{Tabular foundation models for survival analysis.}}
Recent studies have begun to explore the application of foundation models to survival analysis. One line of work reformulates survival prediction as a sequence of binary classification tasks over discretized time intervals, enabling the use of existing tabular models within a standard supervised learning framework~\cite{kim2026}. Other approaches, such as Survival In-Context (SIC), extend the prior-fitted paradigm to survival settings through large-scale synthetic pretraining, allowing models to produce survival predictions directly via in-context inference~\cite{seletkov2026survival}. Although promising, these approaches exhibit important limitations. Reformulation-based methods do not explicitly model the underlying time-to-event structure or censoring mechanism, and often incur computational overhead due to temporal expansion. In contrast, prior-fitted approaches require specialized pretraining procedures and their behavior under realistic clinical conditions remains insufficiently understood. As a result, the applicability of tabular foundation models to real-world clinical survival analysis is still not well established.

\paragraph{Our contribution.}
In contrast, we propose an alternative approach for applying tabular foundation models to clinical survival analysis. We study representative architectures, including TabPFN, TabDPT, and TabICL, and show that with parameter-efficient adaptation, these models can effectively be extended to time-to-event prediction. We demonstrate that this approach achieves competitive performance on survival analysis prediction tasks while maintaining strong robustness across datasets. We highlight its potential as a practical alternative to conventional survival modeling methods.

\section{Methods}
\label{sec:methods}

Tabular FMs are pretrained for standard supervised tasks, whereas survival analysis requires modeling censored time-to-event outcomes with partially observed labels. We study how pretrained FMs can be adapted to clinical survival prediction without modifying backbone weights, under two paradigms: (i) zero-shot in-context learning, and (ii) survival-aware heads training.

\subsection{Preliminary}

\paragraph{\textbf{Problem Setup.}}
In our work, we consider right-censored clinical survival data $\mathcal{D} = \{(\mathbf{x}_i, \tilde{T}_i, \delta_i)\}_{i=1}^N$, where $\mathbf{x}_i \in \mathbb{R}^d$ denotes patient covariates, $T_i$ is the event time of interest, and $C_i$ is the censoring time. Since the event may not be observed for all patients, we observe only the time $\tilde{T}_i = \min(T_i, C_i)$ and an event indicator $\delta_i = \mathbf{1}[T_i \le C_i]$, which specifies whether the event occurred ($\delta_i = 1$) or the observation was censored ($\delta_i = 0$). The objective is to estimate the conditional survival function $S(t \mid \mathbf{x}) = P(T > t \mid \mathbf{x})$, which represents the probability that the event has not occurred by time $t$, or equivalently to learn a risk score that preserves the relative ordering of event times across patients.

\paragraph{\textbf{Problem of reformulation via discrete-time classification.}}
A common strategy is to reformulate survival prediction as a sequence of binary classification problems over ordered time horizons $t_1 < \cdots < t_K$~\cite{kim2026,kvamme2019,yu2011learning}, \highlightsecond{where $K$ is the total number of time bins and $k \in \{1,\ldots,K\}$ indexes individual time bins}. For each patient and time bin, labels are defined as $Y_{ik} = \mathbf{1}[T_i \le t_k]$, and training is performed only on patient--time pairs where the event status is observable. This yields a set of censoring-aware binary tasks from which survival probabilities can be reconstructed as

\begin{equation}
\hat{S}(t_k \mid \mathbf{x}) = 1 - \hat{P}(T \le t_k \mid \mathbf{x}).
\end{equation}
\highlightsecond{Here $\hat{S}{(\cdot)}$ denotes an estimated quantity; specifically, $\hat{P}(T \leq t_k \mid x)$ is the model's predicted event probability at bin $k$.}

While this reformulation enables compatibility with standard classification models, it introduces several limitations. Temporal expansion increases data size by up to a factor of $K$, leading to higher computational cost and class imbalance. The approach is also sensitive to discretization, creating a trade-off between temporal resolution and statistical reliability. Moreover, predictions are learned independently across time, which can produce inconsistencies such as non-monotonic survival curves requiring post-hoc correction. Finally, it does not explicitly model the time-to-event process, limiting its ability to capture temporal dependencies and censoring. These limitations motivate approaches that model survival structure more directly while leveraging pretrained tabular models.

\subsection{Survival Foundation Models}

Let $f_\theta$ denote a pretrained tabular foundation model (e.g., TabPFN). Given a labeled training cohort $\mathcal{D}_{\mathrm{train}}$ and a query patient $\mathbf{x}_i$, the model produces a context-conditioned representation $\mathbf{h}_i = f_\theta(\mathbf{x}_i; \mathcal{D}_{\mathrm{train}})$. This representation captures relationships between the query patient and the training cohort via in-context learning. In all experiments, the backbone $f_\theta$ is kept fixed, and only lightweight prediction heads are trained on top of $\mathbf{h}_i$.

\paragraph{\textbf{Zero-shot in-context learning.}}
We apply the pretrained model directly to the discretized survival task without gradient-based optimization, similar to \cite{kim2026}. For each time bin $t_k$, the model predicts the probability of event occurrence $\hat{p}_k = P(T \le t_k \mid \mathbf{x})$ using the training cohort as in-context support. The survival function is then reconstructed as $\hat{S}(t_k \mid \mathbf{x}) = 1 - \hat{p}_k$ with post-processing to enforce monotonicity across time. This follows the standard discretized reformulation, but leverages in-context learning instead of parameter updates.

\paragraph{\textbf{Survival-aware modelling.}}
To directly model time-to-event structure, we attach a survival-aware head based on multi-task logistic regression (MTLR)~\cite{yu2011learning}. Unlike classification-based reformulations, MTLR models the survival distribution as a \emph{joint probability over valid survival sequences}, capturing dependencies across time rather than treating each time point independently.

Given the representation $\mathbf{h}_i$, the MTLR head produces a sequence of logits $\boldsymbol{\eta}_i = (\eta_{i1}, \dots, \eta_{iK}) = g_\Phi(\mathbf{h}_i)$, where each $\eta_{ik}$ corresponds to a discretized time bin. A survival outcome is encoded as a binary sequence $\mathbf{y} = (y_1,\dots,y_K)$ with the constraint that $\mathbf{y}$ takes the form $(0,\dots,0,1,\dots,1)$, indicating the transition from survival to event. MTLR defines a log-linear model over these valid sequences:

\begin{equation}
P(\mathbf{y} \mid \mathbf{x}_i)
=
\frac{\exp\!\left(\sum_{k=1}^K y_k \, \eta_{ik}\right)}
{\sum_{\mathbf{y}' \in \mathcal{Y}} \exp\!\left(\sum_{k=1}^K y'_k \, \eta_{ik}\right)},
\end{equation}
where $\mathcal{Y}$ denotes the set of all valid survival sequences.

From this distribution, the survival function is obtained by summing over sequences corresponding to survival beyond time $t_k$:
\begin{equation}
S(t_k \mid \mathbf{x}_i) = P(T > t_k \mid \mathbf{x}_i).
\end{equation}

Training is performed by maximizing the likelihood of observed outcomes. For an uncensored observation ($\delta_i = 1$), the likelihood corresponds to the probability of the sequence encoding the observed event time. For a censored observation ($\delta_i = 0$), the likelihood is obtained by marginalizing over all sequences consistent with survival beyond the censoring time.

This formulation directly models a valid survival distribution, enforces monotonicity by construction, and naturally handles censoring without requiring temporal expansion.
\section{Experimental Setup}
\label{sec:experiments}

\paragraph{\textbf{Datasets.}}
We evaluate our approach on two large-scale intensive care unit (ICU) cohorts and a collection of small-medium-scale clinical datasets that represent realistic clinical survival prediction tasks.


\begin{itemize}  
    \item \textbf{eICU}~\cite{pollard2018eicu} is derived from the eICU Collaborative Research Database v2.0 and contains over 200{,}000 ICU admissions across 208 US hospitals. The prediction task is all-cause in-ICU mortality, with survival time measured in hours from ICU admission and censoring defined by discharge alive.
    \item \textbf{MIMIC-IV}~\cite{johnson2023mimic} is constructed from MIMIC-IV v3.1 (2008--2019), restricted to the first patient ICU stay. The task is all-cause in-hospital mortality, with survival time measured from ICU admission to hospital discharge or death.

    \item In addition, we include standard public survival benchmarks, \textbf{SUPPORT}, \textbf{METABRIC}, \textbf{GBSG}, \textbf{WHAS}, \textbf{FLCHAIN}, \textbf{SEER}, and \textbf{VETERANS}, spanning multiple clinical domains (oncology, cardiology, and critical care). 
\end{itemize}

These datasets vary substantially in sample size, feature dimensionality, and censoring rates, providing a complementary testbed to evaluate generalization across data regimes. For the ICU datasets, features are extracted from the first 24 hours of ICU admission to reflect an early-risk prediction setting. The feature sets include demographics, vital signs, laboratory measurements, treatment indicators, and diagnosis information available within the prediction window. Full dataset statistics are provided in Table~\ref{tab:datasets_all}.

\begin{table}[ht]
\centering
\footnotesize
\setlength{\tabcolsep}{5pt}
\renewcommand{\arraystretch}{0.9}
\caption{\textbf{Dataset characteristics.} $N$: number of subjects. Feat.: Number of clinical variables (features). Event rate: proportion of uncensored observations. Med.\,FU: median follow-up time in dataset-native units (IQR in parentheses). Miss.\,\%: mean fraction of missing values across all feature columns. Bins: adaptive time-discretisation intervals for discrete-time heads. EHR datasets (eICU, MIMIC-IV) use a static 24\,h feature snapshot; diagnosis features are time-filtered to prevent discharge-diagnosis leakage.} \label{tab:datasets_all}
\resizebox{0.8\columnwidth}{!}{
\begin{tabular}{lrrrrrr}
\toprule
\textbf{Dataset} & \textbf{$N$} & \textbf{Feat.} & \textbf{Event rate} & \textbf{Med.\,FU (IQR)} & \textbf{Miss.\,\%} & \textbf{Bins}  \\
\midrule
\multicolumn{7}{c}{\textit{(a) Small-medium-scale EHR Benchmarks}} \\
\midrule
Veterans & 137 & 8 & 0.93 & 80 (119) & 0.0 & 20  \\
WHAS500 & 500 & 14 & 0.43  & 632 (1067) & 0.0 & 20 \\
METABRIC & 1,904 & 9 & 0.58  & 115 (124) & 0.0 & 100  \\
GBSG & 2,232 & 7 & 0.57  & 40 (52) & 0.0 & 100 \\
FLCHAIN & 6,524 & 38 & 0.30  & 4303 (1864) & 0.0 & 100  \\
SUPPORT2 & 8,873 & 14 & 0.68  & 231 (737) & 0.0 & 30 \\
SEER & 4,024 & 23 & 0.15  & 73 (34) & 0.0 & 30 \\
\midrule
\multicolumn{7}{c}{\textit{(b) Large-scale EHR Public Benchmarks}} \\
\midrule
eICU & 66,719 & 97 & 0.07  & 88 (76) & 26.6 & 100 \\
MIMIC-IV & 58,329 & 95 & 0.08  & 155 (166) & 16.3 & 100 \\
\bottomrule
\end{tabular}
}
\end{table}

\paragraph{\textbf{Data preprocessing.}}
All features are preprocessed using a consistent pipeline across datasets. Missing values are handled using median imputation, where imputation statistics are computed on the training split and applied to the test split to avoid data leakage. To mitigate the effect of extreme outliers commonly present in clinical measurements, we apply quantile-based clipping (Winsorization) to continuous features. Specifically, values are clipped to the $[0.1\%, 99.9\%]$ quantile range estimated from the training data. \highlightfirst{This step is intended to reduce the influence of measurement errors, data entry artifacts, and implausible values, which are known to occur in electronic health records. Importantly, the chosen thresholds are conservative and affect only a very small fraction of samples, preserving the vast majority of clinically meaningful variation. We emphasize that this procedure does not remove observations, but rather limits the influence of extreme values on model training}. Finally, all numerical features are standardized using z-score normalization based on the training split, ensuring zero mean and unit variance. The same transformation is applied to the test data using the fitted parameters. For discrete-time methods, the number of time bins is selected adaptively from the training data by scanning a set of candidate values and choosing the largest bin count such that each bin contains at least a minimum number of observed events (e.g., 5). This ensures sufficient temporal granularity while avoiding sparsity in low-event regions. The selected number of bins for each dataset is reported in Table~\ref{tab:datasets_all}.

\paragraph{\textbf{Models and Baselines.}}
We evaluate three tabular foundation models—TabPFN, TabDPT, and TabICL—under the two settings described in Section~\ref{sec:methods}: zero-shot in-context learning and survival-aware finetuning using MTLR. In all cases, the pretrained backbone is fixed, and only lightweight prediction heads are trained when applicable. We compare these approaches against a set of strong baselines commonly used in clinical survival analysis \cite{gomez2025benchmarking}. These include classical statistical models (Cox PH), tree-based methods (Random Survival Forests and Gradient Boosting), and neural survival models (DeepSurv, MTLR, DeepHit, DySurv~\cite{mesinovic2026dysurv}, and SurvTRACE~\cite{wang2022survtrace}). Together, these baselines cover a broad range of modeling assumptions and provide competitive references for tabular clinical survival prediction. \highlightsecond{We note that Survival In-Context~\cite{seletkov2026survival} was excluded 
as it is a preprint without a public implementation, precluding reproducible evaluation.}

\paragraph{\textbf{Evaluation Protocol and Metrics.}}
We evaluate models using 5-fold cross-validation, stratified by event indicator and discretized event-time quantiles to preserve the censoring distribution across folds. All models are trained and evaluated on identical splits with consistent preprocessing. As the primary metric, we report the time-dependent concordance index (C-index)~\cite{antolini2005time}, which measures the agreement between predicted and observed event-time ordering. 




\section{Results}
\label{sec:results}

\paragraph{\textbf{Overall performance.}}
Table~\ref{tab:survpfn:summary:c_td} shows the performance of all methods, indicating that \textit{Survival-aware Adaptation} is the most consistent adaptation strategy across datasets. Head-training FMs achieve the best performance on 7 of the 9 benchmarks: TabICL-FT-MTLR attains the highest $C_{\text{td}}$ on METABRIC ($0.664$), WHAS ($0.783$), FLCHAIN ($0.928$), and eICU ($0.797$), while TabDPT-FT-MTLR is best on SEER ($0.737$) and ties for the best result on MIMIC-IV ($0.856$). On the remaining datasets, the best methods are strong task-specific baselines, namely MLP-DeepHit on SUPPORT2 ($0.640$) and DySurv on GBSG ($0.675$), while TabPFN-FT-MTLR achieves the best result on VETERANS ($0.745$). Together, these results show that directly training survival heads on top of pretrained tabular representations yields the strongest and most stable performance overall.

\begin{table}[t]
\centering
\caption{Summary Concordance Index (C-index) results (mean~$\pm$~std, 5-fold CV). \textbf{Bold} = best, \underline{underline} = second-best. $^{\ast}$~indicates statistically significant improvement over the best non-FM baseline ($p<0.05$, Wilcoxon signed-rank test).}
\label{tab:survpfn:summary:c_td}
\resizebox{\columnwidth}{!}{
\begin{tabular}{lcccccccccc}
\toprule
\multirow{2}{*}{\textbf{Model}} & \multirow{2}{*}{\textbf{Rank}} & \multicolumn{7}{c}{\textbf{Public Benchmarks}} & \multicolumn{2}{c}{\textbf{EHR (ICU)}} \\
\cmidrule(lr){3-9} \cmidrule(lr){10-11}
& & \textbf{SUP2} $\uparrow$ & \textbf{METAB} $\uparrow$ & \textbf{GBSG} $\uparrow$ & \textbf{WHAS} $\uparrow$ & \textbf{VET} $\uparrow$ & \textbf{FLC} $\uparrow$ & \textbf{SEER} $\uparrow$ & \textbf{eICU} $\uparrow$ & \textbf{MIMIC-IV} $\uparrow$ \\
\midrule
\multicolumn{11}{l}{\textit{\footnotesize Baselines}} \\
Cox PH & 12.78 & \ensuremath{0.553_{\textcolor{gray}{\pm 0.013}}} & \ensuremath{0.614_{\textcolor{gray}{\pm 0.008}}} & \ensuremath{0.643_{\textcolor{gray}{\pm 0.018}}} & \ensuremath{0.752^{\ast}_{\textcolor{gray}{\pm 0.029}}} & \ensuremath{0.697_{\textcolor{gray}{\pm 0.023}}} & \ensuremath{0.899_{\textcolor{gray}{\pm 0.003}}} & \ensuremath{0.714^{\ast}_{\textcolor{gray}{\pm 0.019}}} & \ensuremath{0.754_{\textcolor{gray}{\pm 0.007}}} & \ensuremath{0.795_{\textcolor{gray}{\pm 0.003}}} \\
RSF & 8.89 & \ensuremath{0.620_{\textcolor{gray}{\pm 0.009}}} & \ensuremath{0.646_{\textcolor{gray}{\pm 0.034}}} & \ensuremath{0.667_{\textcolor{gray}{\pm 0.014}}} & \ensuremath{\underline{0.778}_{\textcolor{gray}{\pm 0.030}}} & \ensuremath{0.698_{\textcolor{gray}{\pm 0.038}}} & \ensuremath{0.913_{\textcolor{gray}{\pm 0.004}}} & \ensuremath{0.720_{\textcolor{gray}{\pm 0.021}}} & \ensuremath{0.715_{\textcolor{gray}{\pm 0.018}}} & \ensuremath{0.807_{\textcolor{gray}{\pm 0.014}}} \\
GBSA & 8.56 & \ensuremath{0.619^{\ast}_{\textcolor{gray}{\pm 0.005}}} & \ensuremath{0.623_{\textcolor{gray}{\pm 0.023}}} & \ensuremath{0.671_{\textcolor{gray}{\pm 0.018}}} & \ensuremath{0.758_{\textcolor{gray}{\pm 0.033}}} & \ensuremath{0.685_{\textcolor{gray}{\pm 0.077}}} & \ensuremath{0.926^{\ast}_{\textcolor{gray}{\pm 0.003}}} & \ensuremath{0.720_{\textcolor{gray}{\pm 0.012}}} & \ensuremath{0.723_{\textcolor{gray}{\pm 0.019}}} & \ensuremath{0.804_{\textcolor{gray}{\pm 0.008}}} \\
\midrule
\multicolumn{11}{l}{\textit{\footnotesize Deep Models}} \\
DeepSurv & 7.00 & \ensuremath{0.611^{\ast}_{\textcolor{gray}{\pm 0.010}}} & \ensuremath{0.638_{\textcolor{gray}{\pm 0.015}}} & \ensuremath{0.670_{\textcolor{gray}{\pm 0.010}}} & \ensuremath{0.757_{\textcolor{gray}{\pm 0.016}}} & \ensuremath{0.657_{\textcolor{gray}{\pm 0.036}}} & \ensuremath{\underline{0.927}_{\textcolor{gray}{\pm 0.004}}} & \ensuremath{0.725_{\textcolor{gray}{\pm 0.015}}} & \ensuremath{0.784^{\ast}_{\textcolor{gray}{\pm 0.008}}} & \ensuremath{\underline{0.844}^{\ast}_{\textcolor{gray}{\pm 0.012}}} \\
MLP-MTLR & 10.33 & \ensuremath{0.625_{\textcolor{gray}{\pm 0.017}}} & \ensuremath{0.625_{\textcolor{gray}{\pm 0.046}}} & \ensuremath{0.659_{\textcolor{gray}{\pm 0.022}}} & \ensuremath{0.720_{\textcolor{gray}{\pm 0.092}}} & \ensuremath{0.653_{\textcolor{gray}{\pm 0.137}}} & \ensuremath{0.924^{\ast}_{\textcolor{gray}{\pm 0.003}}} & \ensuremath{0.681_{\textcolor{gray}{\pm 0.046}}} & \ensuremath{0.765_{\textcolor{gray}{\pm 0.017}}} & \ensuremath{0.827_{\textcolor{gray}{\pm 0.006}}} \\
MLP-DeepHit & 8.33 & \ensuremath{\textbf{0.640}^{\ast}_{\textcolor{gray}{\pm 0.014}}} & \ensuremath{0.654_{\textcolor{gray}{\pm 0.034}}} & \ensuremath{0.662_{\textcolor{gray}{\pm 0.009}}} & \ensuremath{0.716_{\textcolor{gray}{\pm 0.058}}} & \ensuremath{0.675_{\textcolor{gray}{\pm 0.082}}} & \ensuremath{0.923^{\ast}_{\textcolor{gray}{\pm 0.003}}} & \ensuremath{0.688_{\textcolor{gray}{\pm 0.044}}} & \ensuremath{0.777_{\textcolor{gray}{\pm 0.009}}} & \ensuremath{0.834^{\ast}_{\textcolor{gray}{\pm 0.008}}} \\
DySurv & 8.89 & \ensuremath{0.609^{\ast}_{\textcolor{gray}{\pm 0.009}}} & \ensuremath{0.633^{\ast}_{\textcolor{gray}{\pm 0.026}}} & \ensuremath{\textbf{0.675}_{\textcolor{gray}{\pm 0.020}}} & \ensuremath{0.751_{\textcolor{gray}{\pm 0.025}}} & \ensuremath{0.612_{\textcolor{gray}{\pm 0.089}}} & \ensuremath{0.918_{\textcolor{gray}{\pm 0.006}}} & \ensuremath{0.727_{\textcolor{gray}{\pm 0.013}}} & \ensuremath{0.763_{\textcolor{gray}{\pm 0.015}}} & \ensuremath{0.835^{\ast}_{\textcolor{gray}{\pm 0.011}}} \\
SurvTRACE & 12.22 & \ensuremath{0.563_{\textcolor{gray}{\pm 0.016}}} & \ensuremath{0.606_{\textcolor{gray}{\pm 0.030}}} & \ensuremath{0.658_{\textcolor{gray}{\pm 0.014}}} & \ensuremath{0.662_{\textcolor{gray}{\pm 0.033}}} & \ensuremath{0.622_{\textcolor{gray}{\pm 0.112}}} & \ensuremath{0.918_{\textcolor{gray}{\pm 0.005}}} & \ensuremath{0.732_{\textcolor{gray}{\pm 0.017}}} & \ensuremath{0.781_{\textcolor{gray}{\pm 0.008}}} & \ensuremath{0.780_{\textcolor{gray}{\pm 0.020}}} \\
\midrule
\multicolumn{11}{l}{\textit{\footnotesize Zero-Shot}} \\
TabPFN-ZS & 8.11 & \ensuremath{0.623_{\textcolor{gray}{\pm 0.007}}} & \ensuremath{\underline{0.659}_{\textcolor{gray}{\pm 0.017}}} & \ensuremath{0.670_{\textcolor{gray}{\pm 0.009}}} & \ensuremath{0.671_{\textcolor{gray}{\pm 0.054}}} & \ensuremath{0.707_{\textcolor{gray}{\pm 0.049}}} & \ensuremath{0.913_{\textcolor{gray}{\pm 0.007}}} & \ensuremath{0.722_{\textcolor{gray}{\pm 0.012}}} & \ensuremath{0.749_{\textcolor{gray}{\pm 0.012}}} & \ensuremath{0.801_{\textcolor{gray}{\pm 0.011}}} \\
TabDPT-ZS & 13.00 & \ensuremath{0.612^{\ast}_{\textcolor{gray}{\pm 0.008}}} & \ensuremath{0.638_{\textcolor{gray}{\pm 0.015}}} & \ensuremath{0.664_{\textcolor{gray}{\pm 0.012}}} & \ensuremath{0.661_{\textcolor{gray}{\pm 0.067}}} & \ensuremath{0.676_{\textcolor{gray}{\pm 0.065}}} & \ensuremath{0.885_{\textcolor{gray}{\pm 0.013}}} & \ensuremath{0.714_{\textcolor{gray}{\pm 0.031}}} & \ensuremath{0.721_{\textcolor{gray}{\pm 0.011}}} & \ensuremath{0.777_{\textcolor{gray}{\pm 0.007}}} \\
TabICL-ZS & 9.67 & \ensuremath{0.618^{\ast}_{\textcolor{gray}{\pm 0.008}}} & \ensuremath{0.651_{\textcolor{gray}{\pm 0.021}}} & \ensuremath{\underline{0.672}_{\textcolor{gray}{\pm 0.010}}} & \ensuremath{0.681_{\textcolor{gray}{\pm 0.064}}} & \ensuremath{0.670_{\textcolor{gray}{\pm 0.051}}} & \ensuremath{0.911_{\textcolor{gray}{\pm 0.007}}} & \ensuremath{0.724_{\textcolor{gray}{\pm 0.016}}} & \ensuremath{0.747_{\textcolor{gray}{\pm 0.009}}} & \ensuremath{0.802_{\textcolor{gray}{\pm 0.013}}} \\
\midrule
\multicolumn{11}{l}{\textit{\footnotesize Survival-aware Adaptation}} \\
TabPFN-MTLR & 6.11 & \ensuremath{0.610^{\ast}_{\textcolor{gray}{\pm 0.011}}} & \ensuremath{0.655^{\ast}_{\textcolor{gray}{\pm 0.038}}} & \ensuremath{0.662_{\textcolor{gray}{\pm 0.006}}} & \ensuremath{0.760_{\textcolor{gray}{\pm 0.056}}} & \ensuremath{\textbf{0.745}_{\textcolor{gray}{\pm 0.060}}} & \ensuremath{0.924^{\ast}_{\textcolor{gray}{\pm 0.003}}} & \ensuremath{0.728_{\textcolor{gray}{\pm 0.018}}} & \ensuremath{0.781_{\textcolor{gray}{\pm 0.006}}} & \ensuremath{0.841^{\ast}_{\textcolor{gray}{\pm 0.006}}} \\
TabDPT-MTLR & \underline{5.11} & \ensuremath{0.625_{\textcolor{gray}{\pm 0.012}}} & \ensuremath{0.653_{\textcolor{gray}{\pm 0.038}}} & \ensuremath{0.669_{\textcolor{gray}{\pm 0.017}}} & \ensuremath{0.775_{\textcolor{gray}{\pm 0.035}}} & \ensuremath{0.703_{\textcolor{gray}{\pm 0.091}}} & \ensuremath{0.886_{\textcolor{gray}{\pm 0.005}}} & \ensuremath{\textbf{0.737}_{\textcolor{gray}{\pm 0.008}}} & \ensuremath{\underline{0.794}^{\ast}_{\textcolor{gray}{\pm 0.008}}} & \ensuremath{\textbf{0.856}^{\ast}_{\textcolor{gray}{\pm 0.007}}} \\
TabICL-MTLR & \textbf{2.22} & \ensuremath{\underline{0.630}_{\textcolor{gray}{\pm 0.013}}} & \ensuremath{\textbf{0.664}^{\ast}_{\textcolor{gray}{\pm 0.027}}} & \ensuremath{0.670_{\textcolor{gray}{\pm 0.021}}} & \ensuremath{\textbf{0.783}_{\textcolor{gray}{\pm 0.025}}} & \ensuremath{\underline{0.712}_{\textcolor{gray}{\pm 0.059}}} & \ensuremath{\textbf{0.928}_{\textcolor{gray}{\pm 0.002}}} & \ensuremath{\underline{0.736}_{\textcolor{gray}{\pm 0.012}}} & \ensuremath{\textbf{0.797}^{\ast}_{\textcolor{gray}{\pm 0.008}}} & \ensuremath{\textbf{0.856}^{\ast}_{\textcolor{gray}{\pm 0.006}}} \\
\bottomrule
\end{tabular}
}
\end{table}

\paragraph{\textbf{Comparison with zero-shot inference.}}
Zero-shot FM inference is competitive, especially in small-medium-scale settings, but is consistently weaker than survival-aware adaptation. For example, on eICU the best zero-shot method reaches $0.749$ (TabPFN-ZS), compared with $0.797$ for TabICL-FT-MTLR; on MIMIC-IV, the best zero-shot score is $0.802$ (TabICL-ZS), versus $0.856$ for fine-tuned survival heads. Similar improvements appear on FLCHAIN ($0.913$ for TabPFN-ZS vs.\ $0.928$ for TabICL-FT-MTLR) and VETERANS ($0.707$ for TabPFN-ZS vs.\ $0.745$ for TabPFN-FT-MTLR). This pattern is consistent with the limitations of zero-shot discretized reformulation: each patient must be expanded into multiple patient--time pairs, increasing computational cost and class imbalance, while the model still predicts bin-wise outcomes independently. In contrast, survival-aware adaptation operates directly at the patient level and better captures the joint time-to-event structure.

\paragraph{\textbf{Effect of pretrained backbone representations.}}
The results also highlight the value of foundation-model representations over standard MLP encoders. Comparing MTLR heads with the same survival objective, TabPFN-FT-MTLR outperforms MLP-MTLR on 7 of 9 datasets, including large gains on VETERANS ($0.745$ vs.\ $0.653$), METABRIC ($0.655$ vs.\ $0.625$), and MIMIC-IV ($0.841$ vs.\ $0.827$). Likewise, TabICL-FT-MTLR substantially improves over MLP-MTLR on WHAS ($0.783$ vs.\ $0.720$), FLCHAIN ($0.928$ vs.\ $0.924$), eICU ($0.797$ vs.\ $0.765$), and MIMIC-IV ($0.856$ vs.\ $0.827$). Even against the stronger MLP-DeepHit baseline, FM-based survival heads remain competitive or superior on several datasets, particularly on WHAS, VETERANS, FLCHAIN, eICU, and MIMIC-IV. These comparisons suggest that pretrained tabular backbones, especially TabPFN-style models, provide richer representations than task-specific MLP encoders and transfer effectively to censored survival prediction.

\paragraph{\textbf{Performance on large-scale EHR data.}}
The benefits of survival-aware adaptation persist on large ICU cohorts. On eICU, TabICL-FT-MTLR achieves the best result ($0.797$), improving over Cox PH ($0.777$), DeepSurv ($0.784$), SurvTRACE ($0.781$), and the best zero-shot FM ($0.749$). On MIMIC-IV, both TabDPT-FT-MTLR and TabICL-FT-MTLR reach the top performance ($0.856$), outperforming Cox PH ($0.820$), DeepSurv ($0.845$), MLP-DeepHit ($0.834$), and zero-shot inference ($0.801$--$0.802$). These results indicate that the gains from pretrained tabular representations and survival-aware training are not limited to small benchmarks, but extend to large, heterogeneous EHR datasets as well.

\paragraph{\textbf{Risk stratification analysis.}}

\begin{figure}
\centering

\textbf{A. TabPFN} \hfill \textbf{B. TabDPT} \hfill \textbf{C. TabICL} \\
\begin{subfigure}[t]{0.32\linewidth}
    \includegraphics[width=\linewidth]{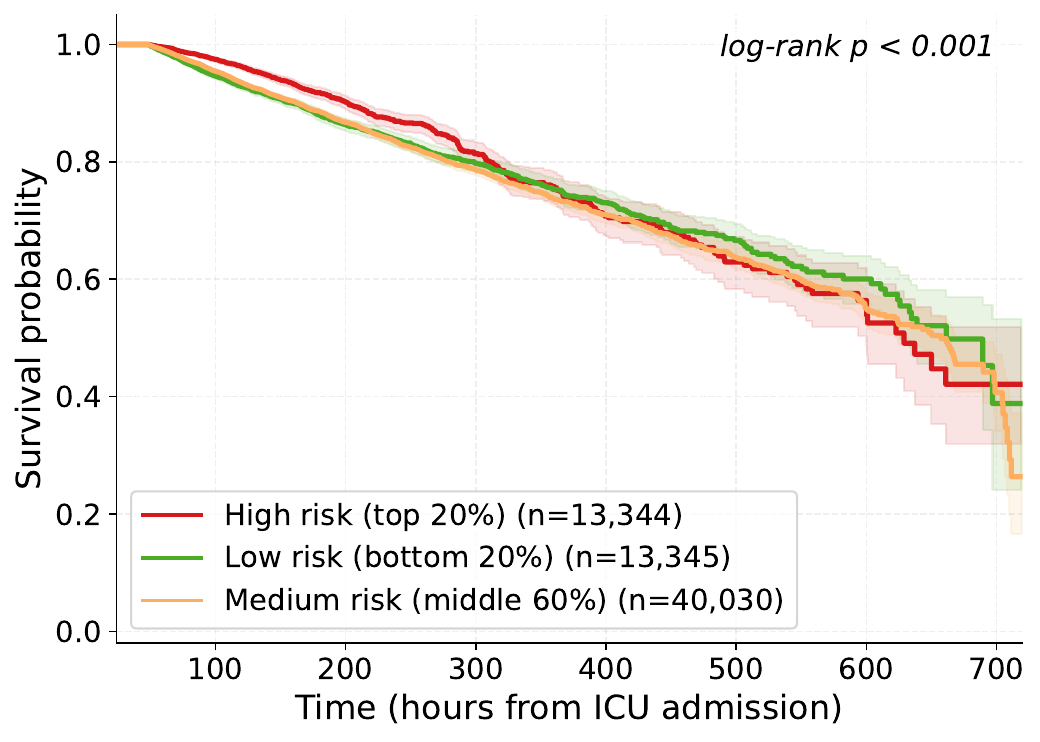}
\end{subfigure}
\hfill
\begin{subfigure}[t]{0.32\linewidth}
    \includegraphics[width=\linewidth]{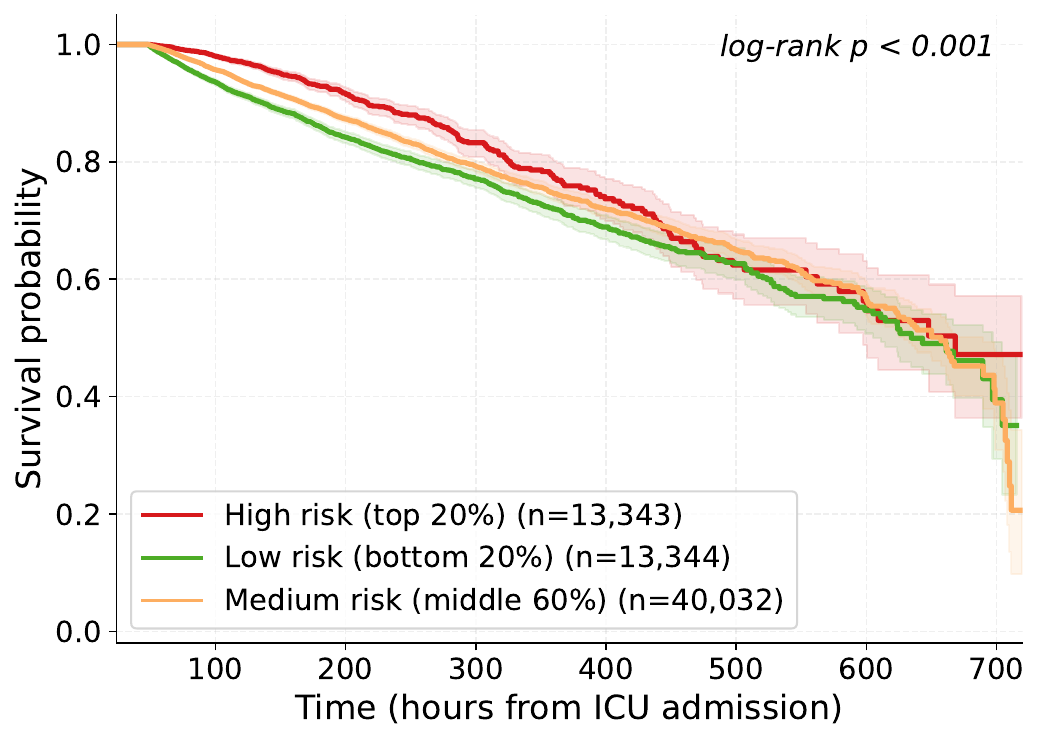}
\end{subfigure}
\hfill
\begin{subfigure}[t]{0.32\linewidth}
    \includegraphics[width=\linewidth]{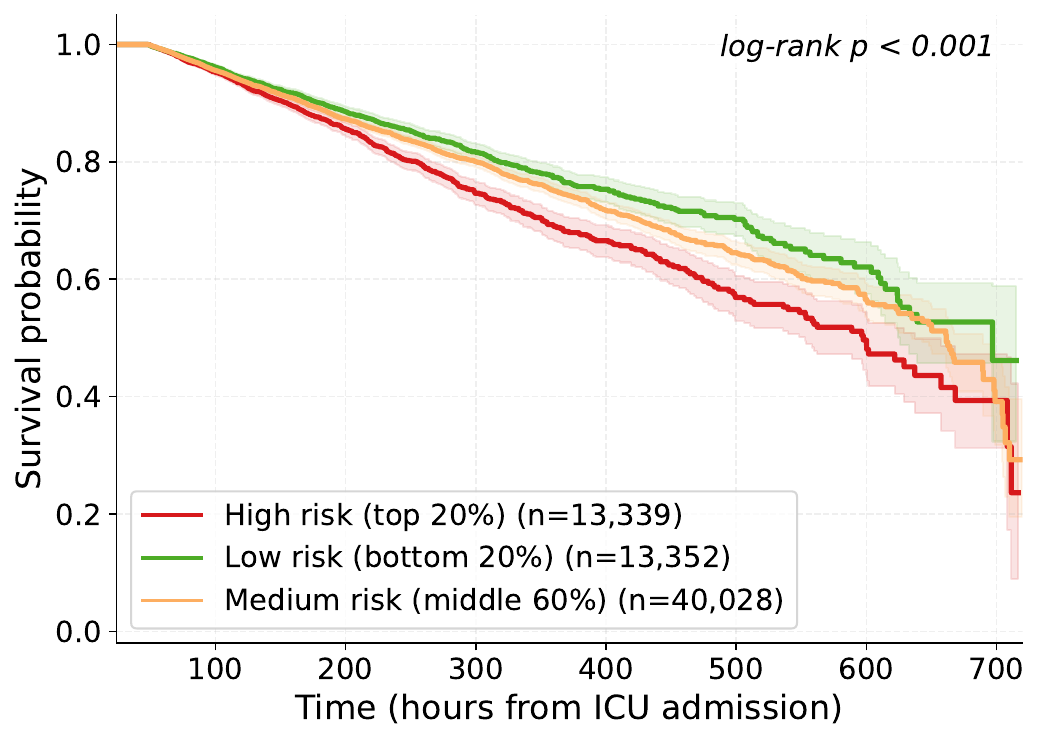}
\end{subfigure}

\vspace{0.3em}
\textit{(a) eICU — Zero-shot}

\vspace{0.4em}

\begin{subfigure}[t]{0.32\linewidth}
    \includegraphics[width=\linewidth]{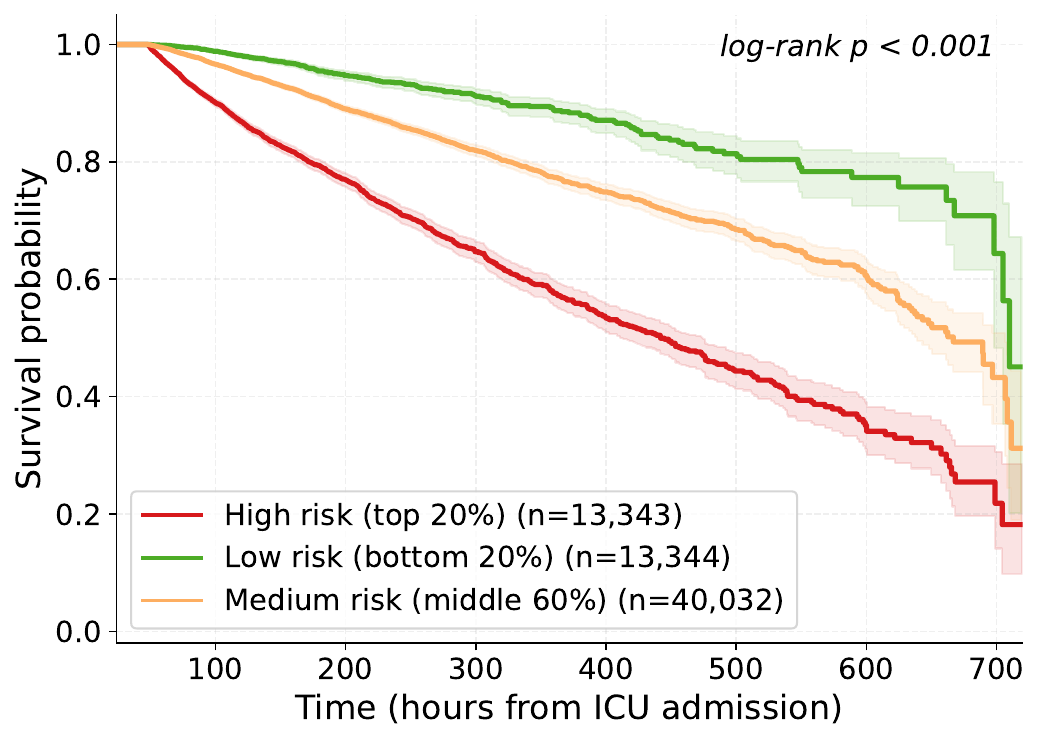}
\end{subfigure}
\hfill
\begin{subfigure}[t]{0.32\linewidth}
    \includegraphics[width=\linewidth]{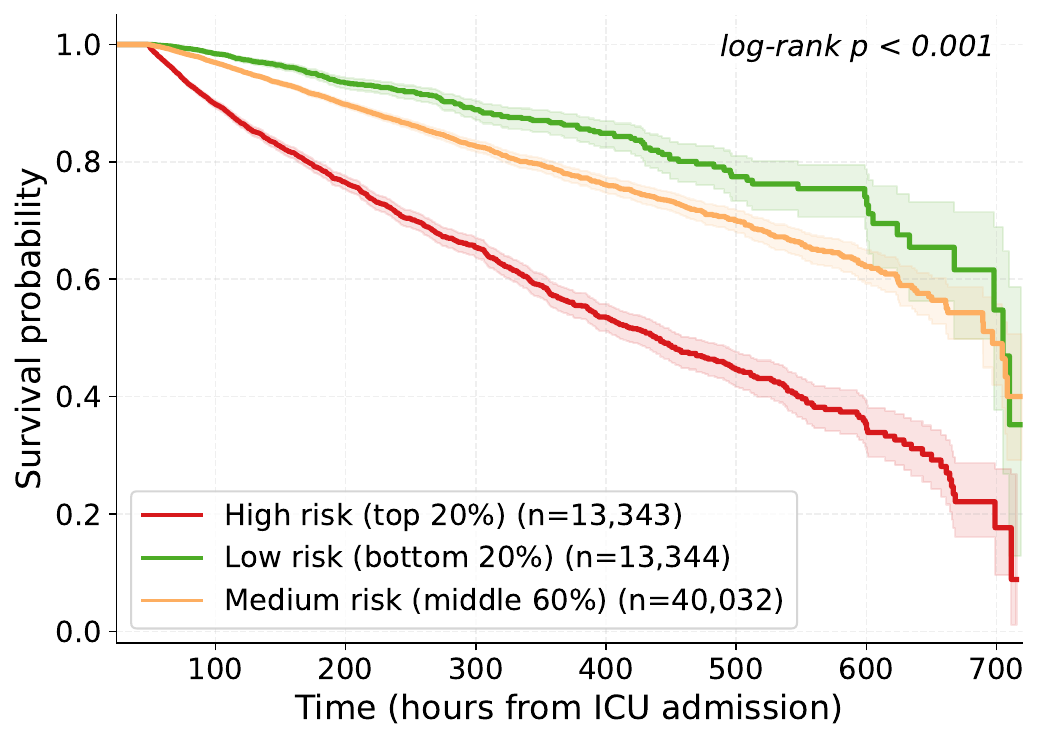}
\end{subfigure}
\hfill
\begin{subfigure}[t]{0.32\linewidth}
    \includegraphics[width=\linewidth]{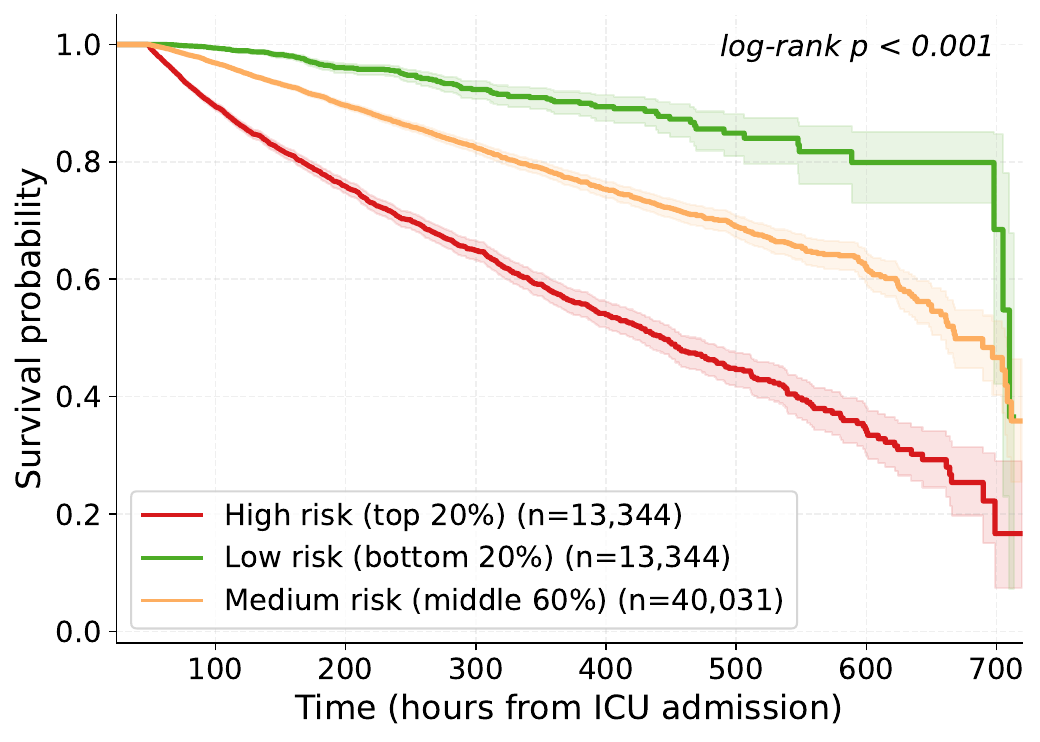}
\end{subfigure}

\vspace{0.3em}
\textit{(b) eICU — Survival-aware Adaptation}

\vspace{0.6em}

\begin{subfigure}[t]{0.32\linewidth}
    \includegraphics[width=\linewidth]{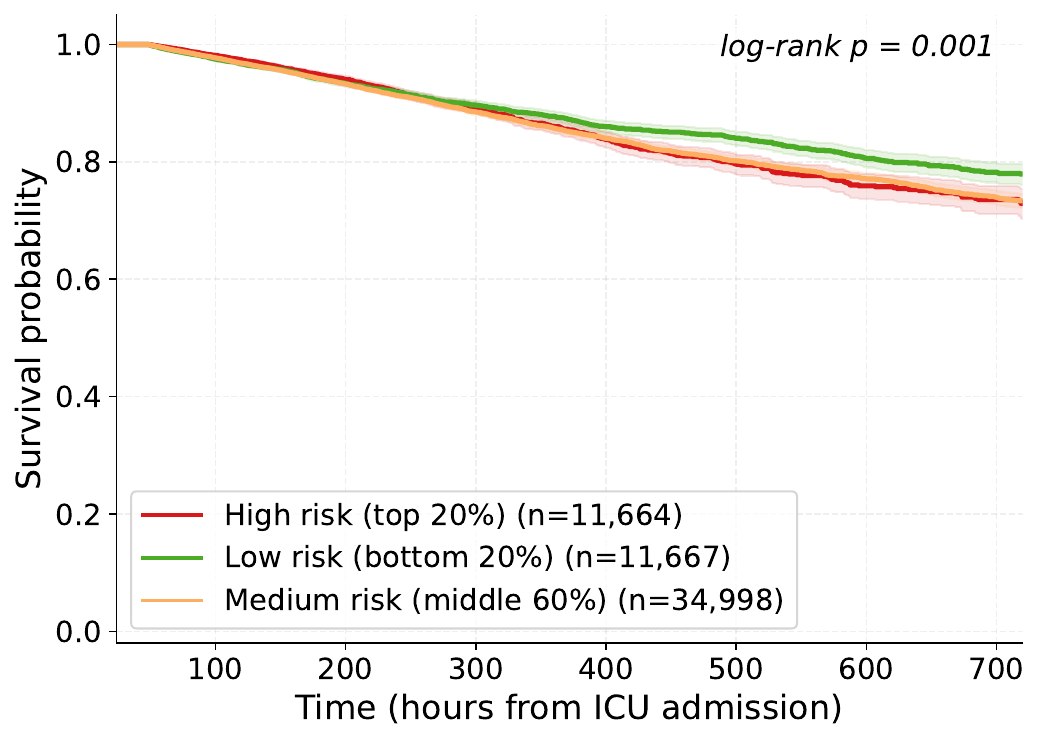}
\end{subfigure}
\hfill
\begin{subfigure}[t]{0.32\linewidth}
    \includegraphics[width=\linewidth]{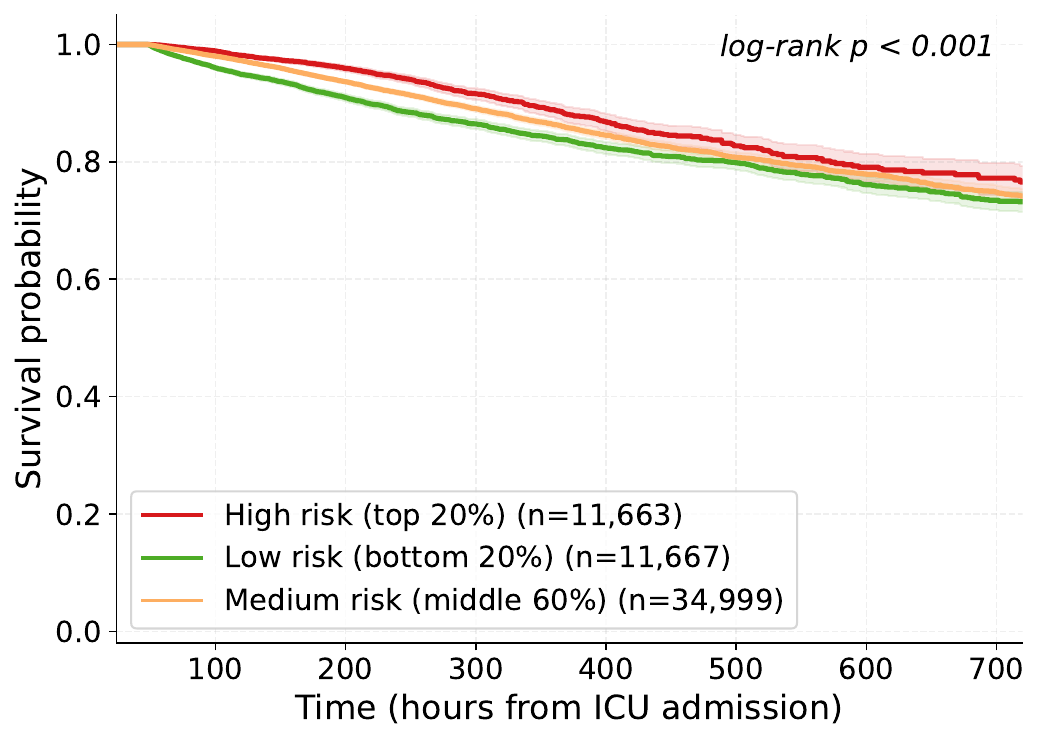}
\end{subfigure}
\hfill
\begin{subfigure}[t]{0.32\linewidth}
    \includegraphics[width=\linewidth]{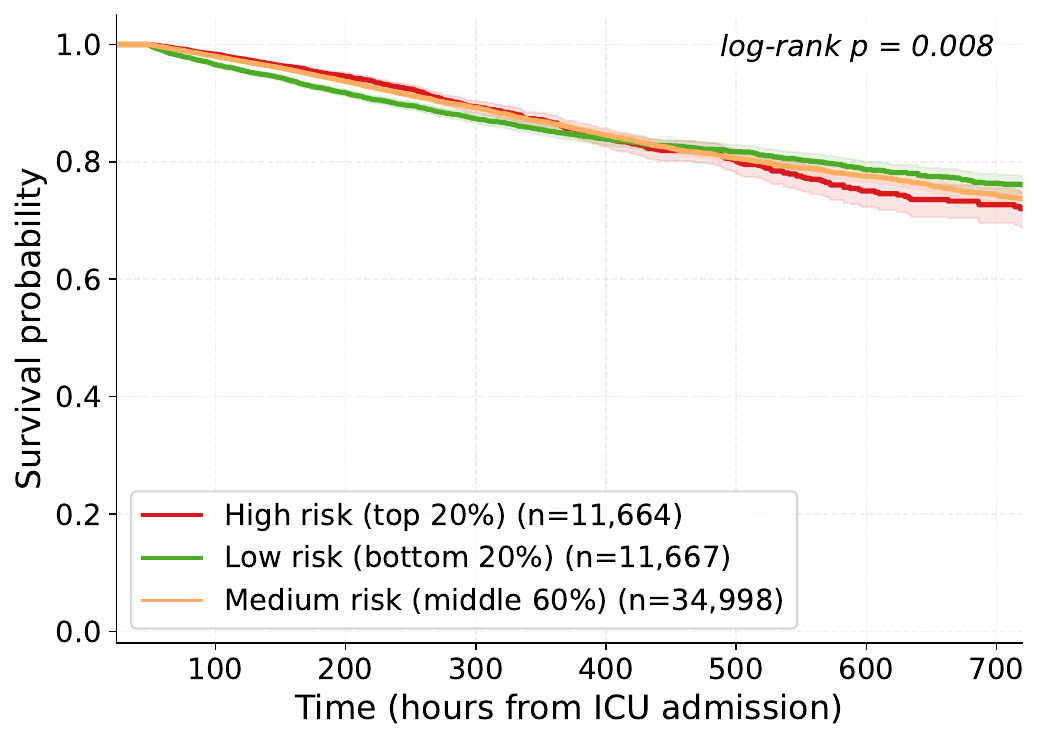}
\end{subfigure}

\vspace{0.3em}
\textit{(c) MIMIC-IV — Zero-shot}

\vspace{0.4em}

\begin{subfigure}[t]{0.32\linewidth}
    \includegraphics[width=\linewidth]{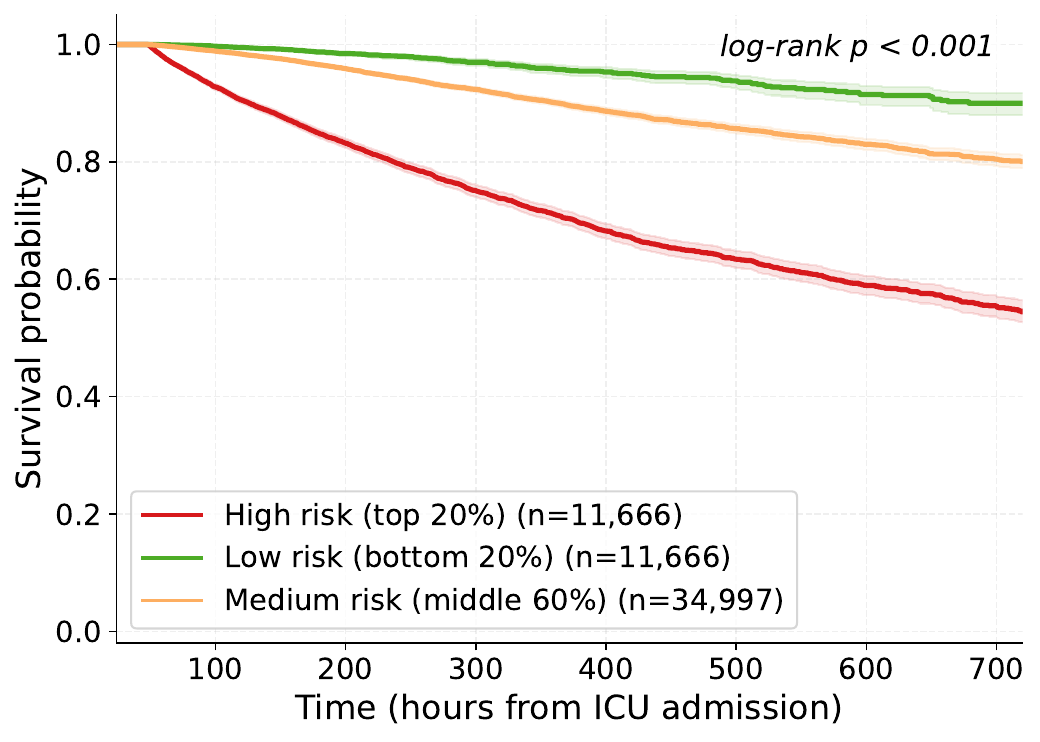}
\end{subfigure}
\hfill
\begin{subfigure}[t]{0.32\linewidth}
    \includegraphics[width=\linewidth]{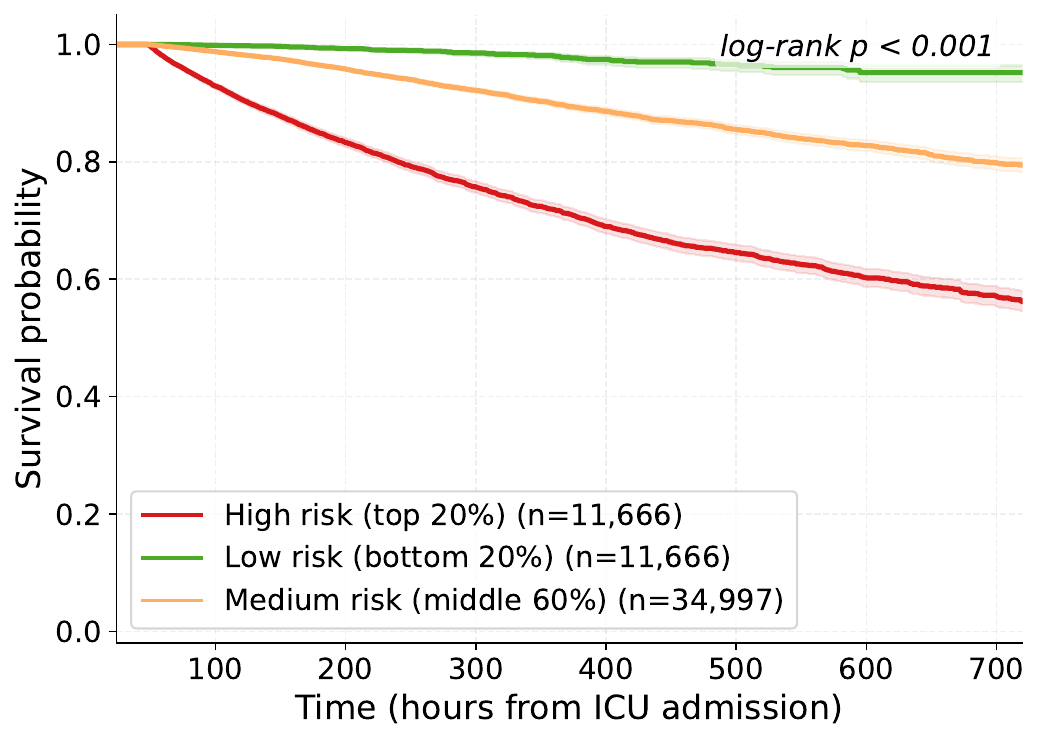}
\end{subfigure}
\hfill
\begin{subfigure}[t]{0.32\linewidth}
    \includegraphics[width=\linewidth]{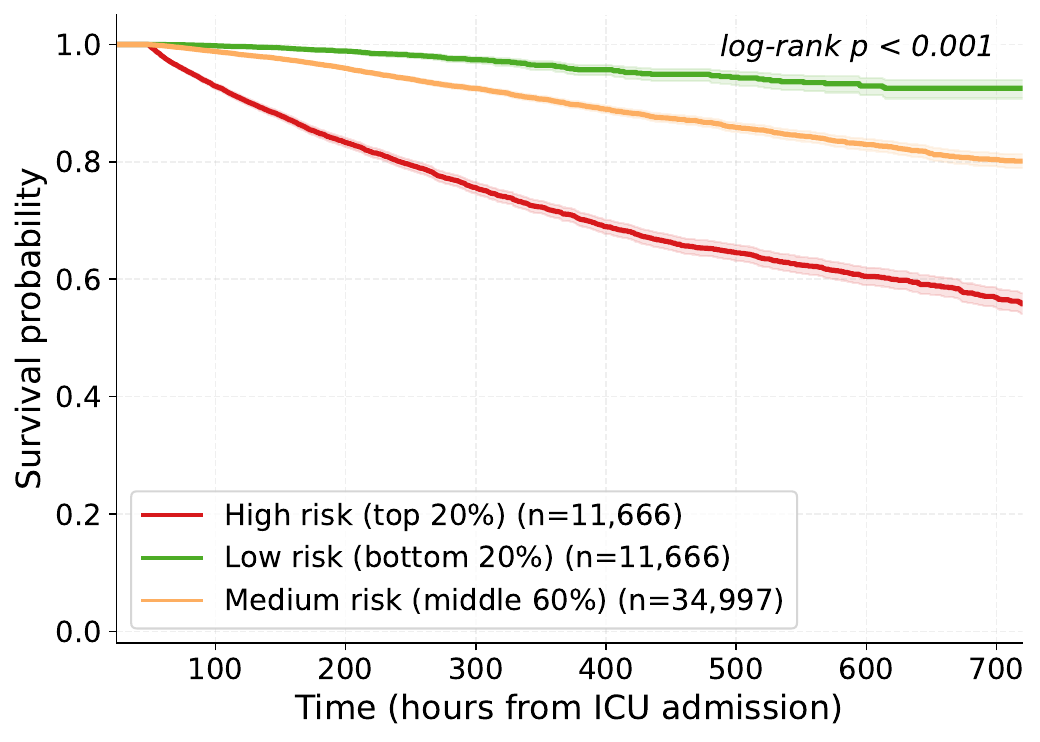}
\end{subfigure}

\vspace{0.3em}
\textit{(d) MIMIC-IV — Survival-aware Adaptation}

\caption{\textbf{Risk-stratified Kaplan--Meier curves across datasets and adaptation strategies.} \highlightsecond{Solid lines show Kaplan--Meier estimates; shaded bands indicate pointwise 95\% confidence intervals. Log-rank $p$-values assess group separation.} Row (a) and (c) show curves for zero-shot foundation models, which exhibit weaker separation and greater overlap between risk groups. Row (b) and (d) show curves for trained survival-aware adapted foundation models, which yield clearer and more stable stratification across both datasets.}
\label{fig:risk_stratification}
\end{figure}

We evaluate clinical utility by grouping patients into low-, medium-, and high-risk cohorts based on model-predicted risk scores (top 20\%, middle 60\% and bottom 20\%). For each group, we compute Kaplan--Meier survival curves over a unified 24--720 hour (30-day) window using the observed outcomes and show in Figure~\ref{fig:risk_stratification}.

Across both eICU and MIMIC-IV, zero-shot foundation models exhibit weaker separation, with noticeable overlap between medium- and high-risk groups, indicating limited resolution in distinguishing patients at elevated risk. While statistical significance is often retained (e.g., log-rank $p = 0.001$--$0.008$ in some settings), the visual separation remains modest. 

In contrast, trained survival-aware models yield clearer and more stable stratification across all architectures and datasets, with well-ordered curves, minimal overlap, and consistently strong statistical significance (log-rank $p < 0.001$). Notably, survival-aware models achieve earlier divergence of risk groups within the first 100--200 hours, suggesting improved identification of patients at imminent risk, which is critical for ICU decision-making. Furthermore, adapted models exhibit fewer curve crossings, indicating more consistent risk ranking over time. 

Overall, these results demonstrate that while zero-shot models provide reasonable ranking performance, survival-aware adaptation substantially improves the separation of clinically distinct risk groups, enabling more reliable and actionable risk stratification in real-world ICU settings.

\section{Discussion}
\label{sec:discussion}

Across datasets, we observe that the benefits of tabular foundation models depend strongly on data complexity and heterogeneity. On large and diverse ICU cohorts such as eICU and MIMIC-IV, survival-aware training mostly provides substantial gains over both classical and deep baselines, suggesting that pretrained representations capture complex feature interactions that are difficult to learn from scratch. In contrast, on more structured datasets such as SUPPORT2 or GBSG, where classical models already perform strongly, the advantage of foundation models is less pronounced. In these settings, zero-shot inference is often sufficient, and additional lightweight adaptation provides only incremental improvements. This highlights that the value of foundation models is most evident in regimes with higher dimensionality, noise, or heterogeneity.

A consistent pattern across experiments is the trade-off between zero-shot inference and survival-aware training. Zero-shot methods are competitive, particularly in smaller or cleaner datasets, and offer the advantage of requiring no training. However, they rely on discretized reformulations and independent bin-wise predictions, which limits their ability to model the joint structure of time-to-event outcomes. In contrast, survival-aware training with MTLR provides a more principled treatment of censoring and temporal dependencies, leading to more consistent gains across datasets. From a practical perspective, this suggests a simple deployment strategy: zero-shot inference can serve as a strong baseline in low-resource settings, while head training becomes increasingly beneficial as dataset size and complexity grow.

Our results also demonstrate that pretrained tabular backbones consistently outperform standard neural encoders when paired with the same survival objective. The improvements observed over MLP-based models indicate that foundation models encode useful inductive biases for tabular data, likely reflecting patterns learned from large collections of synthetic tasks. These representations appear to transfer effectively to survival prediction, even though the pretraining objective does not explicitly involve censoring or time-to-event modeling. This suggests that much of the difficulty in clinical survival analysis lies in representation learning rather than survival-specific loss design.

From a clinical perspective, our findings highlight the potential of tabular foundation models as practical tools for survival prediction. The ability to achieve competitive performance with parameter-efficient adaptation simplifies the modeling pipeline and reduces the need for extensive hyperparameter tuning or architecture design. In particular, the strong performance of zero-shot and lightly fine-tuned models suggests that foundation models may be especially useful in settings where labeled data are limited or rapid deployment is required. Importantly, risk stratification analysis shows that trained survival-aware models produce clearer and more stable separation of clinically meaningful risk groups, while zero-shot models exhibit weaker stratification despite comparable ranking performance. At the same time, the observed improvements under cross-cohort evaluation indicate that these models can generalize reasonably well across institutions, an important consideration for real-world use.

This study has several limitations. First, we focus on static tabular features extracted from early ICU data and do not consider longitudinal or time-varying covariates, which are important in many clinical scenarios. Second, we restrict our analysis to single-event survival and do not explore competing risks or multi-event settings. Third, while we observe strong empirical performance, the mechanisms underlying the transferability of foundation model representations to survival tasks remain unclear. \highlightsecond{Finally, feature-level interpretability of the foundation model representations is missing due to page constraints, and a comparison with SIC~\cite{seletkov2026survival} was not possible due to the lack of a public SIC implementation. We will investigate all these aspects in future work. } Furthermore, we will also explore different survival heads and conduct more principled integration of censoring-aware objectives during pretraining.

\section{Conclusion}
\label{sec:conclusion}

We presented an approach for applying tabular foundation models to clinical survival analysis by adapt the survival-specific heads on pretrained tabular representations. Using representative architectures, we showed that this lightweight adaptation with a multi-task logistic regression head enables effective modeling of censored time-to-event outcomes without modifying backbone parameters.

Across diverse benchmarks and large-scale ICU cohorts, fine-tuned tabular foundation models achieved competitive or superior performance compared to classical statistical, tree-based, and neural survival models. Survival-aware adaptation consistently outperformed zero-shot inference and standard neural baselines, with gains particularly evident in large and heterogeneous datasets such as eICU and MIMIC-IV.

Beyond standard metrics, our results demonstrate that tabular foundation models enable improved clinical risk stratification, producing well-separated and temporally consistent patient groups aligned with observed outcomes. In particular, survival-aware adaptation leads to clearer identification of high-risk patients and earlier separation of risk groups, supporting more timely and actionable decision-making in ICU settings.

Overall, these findings highlight the importance of combining pretrained tabular representations with survival-aware objectives. This approach provides a practical and effective alternative to conventional survival modeling, offering strong predictive performance, clinically meaningful stratification, and a simplified training pipeline. Future work may extend this framework to longitudinal data, competing risks, and tighter integration of survival objectives during pretraining.

\begin{credits}
\subsubsection{\ackname}
This research was conducted with the financial support of Taighde Éireann-Research Ireland under Grant Agreement No. 13/RC/2106\_P2 at ADAPT, the Research Ireland Centre for AI-Driven Digital Content Technology at DCU funded through the Research Ireland Research Centres Programme. 

\subsubsection{\discintname}
The authors have no competing interests to declare.
\end{credits}

\newpage

\bibliographystyle{splncs04}
\bibliography{refs}

\end{document}